\documentclass[conference]{IEEEtran}
\IEEEoverridecommandlockouts
\usepackage{amsmath,amssymb,amsfonts}
\usepackage{algorithmic}
\usepackage{graphicx}
\usepackage{textcomp}
\usepackage{xcolor}
\usepackage{booktabs}
\usepackage{amsthm}
\usepackage{subcaption}
\usepackage{float}

\usepackage[
  backend=biber,
  style=ieee,
  maxbibnames=1,  
  minbibnames=1,  
  maxcitenames=3, 
  mincitenames=3  
]{biblatex}

\def\BibTeX{{\rm B\kern-.05em{\sc i\kern-.025em b}\kern-.08em
    T\kern-.1667em\lower.7ex\hbox{E}\kern-.125emX}}
\begin{document}

\title{Fuzzy Encoding-Decoding to Improve Spiking Q-Learning Performance in Autonomous Driving}

\author{
\IEEEauthorblockN{
Aref Ghoreishee\thanks{Corresponding author: Aref Ghoreishee (ag4247@drexel.edu).},
Abhishek Mishra,
Lifeng Zhou,
John Walsh,
Anup Das,
and Nagarajan Kandasamy
}
\IEEEauthorblockA{
\textit{Electrical and Computer Engineering Department} \\
\textit{Drexel University}\\
Philadelphia, USA \\
}
}

\maketitle

\begin{abstract}
This paper develops an end-to-end fuzzy encoder–decoder architecture for enhancing vision-based multi-modal deep spiking Q-networks in autonomous driving. The method addresses two core limitations of spiking reinforcement learning: information loss stemming from the conversion of dense visual inputs into sparse spike trains, and the limited representational capacity of spike-based value functions, which often yields weakly discriminative Q-value estimates. The encoder introduces trainable fuzzy membership functions to generate expressive, population-based spike representations, and the decoder uses a lightweight neural decoder to reconstruct continuous Q-values from spiking outputs. Experiments on the \texttt{HighwayEnv} benchmark show that the proposed architecture substantially improves decision-making accuracy and closes the performance gap between spiking and non-spiking multi-modal Q-networks. The results highlight the potential of this framework for efficient and real-time autonomous driving with spiking neural networks.
\end{abstract}


\section{Introduction}
Reinforcement learning (RL) enables autonomous agents to acquire complex decision-making strategies through interaction with dynamic and uncertain environments. Deep reinforcement learning (DRL) extends this capability by integrating deep neural networks with RL-based policy optimization, allowing agents to process high-dimensional sensory inputs and learn effective control policies directly from raw data. Among DRL methods, Deep Q-Networks (DQNs) have emerged as a foundational framework for sequential decision-making in complex domains, supporting robust, data-driven behavior without the need for handcrafted decision rules. Through mechanisms such as experience replay, target networks, and stabilized value iteration~\cite{mnih2015human}, DQNs provide the reliability required for real-time autonomous systems, including autonomous driving, robot navigation, and multi-agent coordination~\cite{zhu2017target,ha2018automated,hu2021sim,zhang2021tactical}, where decisions must be executed under uncertainty and partial observability.

Given the strong capability of deep neural networks to extract high-level patterns from visual data, including spatial structures, geometric relationships, and complex semantic cues, modern autonomous systems often rely heavily on vision-based sensors or convert heterogeneous measurements (e.g., LiDAR, radar, IMU) into image-like representations before downstream processing~\cite{li2024bevformer, lang2019pointpillars, philion2020lift}. When combined with advanced perception modules, such as attention-based fusion layers, DQNs offer a powerful paradigm for integrating diverse sensory modalities into a unified state representation, enabling precise and context-aware decision-making. This integration has demonstrated notable success in autonomous driving and robotic navigation. However, the significant computational and energy demands of conventional DQNs, particularly those employing attention-based fusion mechanisms, remain a major barrier to deployment in resource-limited autonomous platforms ~\cite{hu2024high, tang2021deep, ghoreishee2025improving}.

Spiking Neural Networks (SNNs) offer an alternative that is inherently energy-efficient and event-driven. By transmitting information through sparse binary spikes and replacing expensive multiply--accumulate operations with low-cost accumulations, SNNs provide a solution for real-time decision-making on edge-level autonomous platforms. Recent work by Ghoreishee et al. introduced a vision-based multi-modal deep spiking Q-network (MM-DSQN) for high-level autonomous driving decisions, demonstrating a promising direction toward computationally and energy-efficient on-board intelligence~\cite{ghoreishee2025}. Despite these advantages, previous studies have reported a notable performance gap between spiking and non-spiking DQNs for vision-based decision-making tasks~\cite{ghoreishee2025improving, liu2022human,ghoreishee2025}. We hypothesize that two key limitations of SNN-based DQN pipelines contribute to this gap:
\begin{enumerate}
    \item \emph{Information loss in spike encoding:} Converting continuous visual signals into binary spike trains discards fine-grained perceptual information, degrading the quality of the learned state representation and, consequently, the decision-making performance.
    
    \item \emph{Sparse and quantized Q-values:} The event-driven nature of SNN outputs can cause Q-values to collapse into a narrow and weakly discriminative range, reducing the expressiveness of the Q-function and impairing the subsequent action selection.
\end{enumerate}

We propose a computationally efficient \emph{population-based fuzzy encoding and decoding approach} for vision-based deep spiking Q-networks. Inspired by fuzzification and defuzzification principles, our method learns optimal fuzzy membership functions to minimize information loss during spike encoding and introduces a population-based decoder to reconstruct expressive Q-values from sparse spike outputs. To the best of our knowledge, this is the first work to integrate a fuzzy population-based encoder--decoder mechanism within spiking MM-DQNs for autonomous driving. The main contributions of this work can be summarized as follows.
\begin{itemize}
    \item \emph{A new population-based fuzzy encoder} that significantly reduces information loss during visual-to-spike conversion by learning compact and computationally efficient fuzzy membership functions.
    
    \item \emph{A population-based fuzzy decoder} that reconstructs continuous Q-values from event-driven neural activity, mitigating the sparse Q-value problem in SNNs.
\end{itemize} 

Evaluation on the \texttt{HighwayEnv} benchmark ~\cite{highway-env} demonstrates that our approach substantially narrows the performance gap between spiking and non-spiking variants while maintaining computational efficiency. \emph{Although the original MM-DSQN exhibits a performance drop of approximately $12.5\%$ compared to the non-spiking variant, the encoder--decoder solution, when integrated within this MM-DSQN, allows the spiking model to match the performance of its non-spiking counterpart.}

The remainder of the paper is organized as follows. Section~\ref{sec:related_work} discusses related work. Section~\ref{sec:preliminaries} discusses concepts necessary to understand our solution. Section~\ref{sec:approach} develops the encoder--decoder approach, and Section~\ref{sec:analysis} provides a theoretical analysis of its key properties. Section~\ref{sec:evaluation} evaluates its performance. We conclude the paper in Section~\ref{sec:conclusions}. 

\begin{figure*}[!t] 
  \centering
  \includegraphics[width=1.0\linewidth]{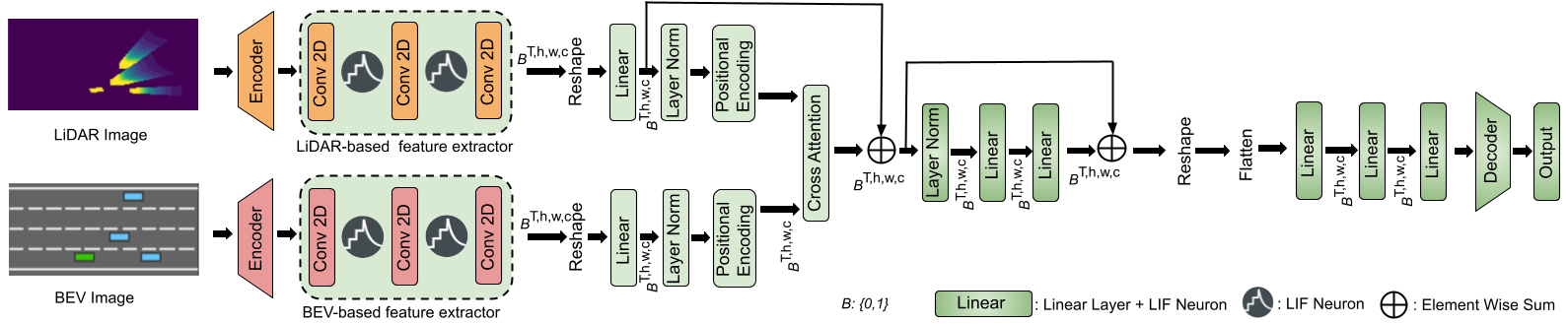}
  \caption{The MM-DSQN architecture that uses a spiking cross-attention module to fuse BEV images and LiDAR representations for RL-based decision making (adapted from Ghoreishee et al.~\cite{ghoreishee2025}).}
  \vspace{-12pt}
  \label{fig1}
\end{figure*}

\section{Related Work}\label{sec:related_work}
Previous research on spiking multimodal decision-making covers two main areas: deep spiking reinforcement learning and multimodal reinforcement learning. Both aim to achieve real-time perception, decision-making, and control under computational and energy constraints, with the goal of enabling efficient on-board intelligence for autonomous systems.

\subsection{Deep Spiking Reinforcement Learning}
The Deep Q-Network (DQN) combines Q-learning with neural networks to enable end-to-end learning from high-dimensional observations~\cite{mnih2013playing}. Techniques such as experience replay and $\epsilon$-greedy exploration~\cite{mnih2015human}, together with Double DQN~\cite{van2016deep}, Dueling DQN~\cite{wang2016dueling}, and double--dueling variants~\cite{sewak2019deep}, improve value estimation and training stability.

To address the high energy cost of DRL, a growing line of work explores deep spiking Q-networks (DSQNs). Early efforts converted pre-trained DQNs into spiking models~\cite{patel2019improved}, followed by more advanced conversion techniques~\cite{tan2021strategy}, direct training with pseudo-gradients \cite{liu2022human}, and spiking-specific batch normalization \cite{sun2022solving}. Despite these advances, many SNN-based RL methods, particularly those in~\cite{sun2022solving,liu2022human,tan2021strategy}, omit explicit spike encoding and instead rely on the first layer to convert continuous inputs into spikes. Although this approach minimizes information loss, it compromises computational efficiency and weakens the event-driven benefits of SNNs.

Population-based spike encoding using radial basis functions has recently been explored for continuous-control RL problems~\cite{jaisankar2025population, tang2021deep}. By increasing the representational capacity of SNNs, these methods demonstrate strong performance on low-dimensional proprioceptive and kinematic observations. However, they do not readily extend to vision-based DSQNs---the main bottleneck being that evaluating exponential kernels for every pixel incurs substantial computational overhead, rendering these encoders inefficient and often infeasible for high-dimensional image inputs that require fast and accurate spike conversion. Thus, there remains a significant gap in developing computationally efficient, scalable encoding and decoding mechanisms suited to visual DSQNs.

\subsection{Multi-Modal Reinforcement Learning}
Multimodal RL combines heterogeneous sensory streams for robust decision-making. For example, Chen et al. maximized shared information between vision, touch, and proprioception sensors using a latent state-space model with a mutual-information objective~\cite{chen2021multi}. Becker et al. used modality-specific self-supervised losses to learn unified state representations from high-dimensional (vision) and low-dimensional (proprioceptive) signals~\cite{becker2023combining}. Jangir et al.\ introduced a transformer-based RL architecture that fuses egocentric and third-person visual streams through cross-view attention~\cite{jangir2022look}. Su et al.\ extended cross-attention fusion to autonomous driving by combining a spatio-temporal agent graph with camera features~\cite{su2024multimodal}, while Lu et al.\ integrated heterogeneous IoT sensor data with contextual information (weather and traffic) for decision-making in global logistics~\cite{lu2025multimodal}.

A recent multimodal deep spiking Q-network incorporating a spiking cross-attention mechanism, proposed by Ghoreishee et al., represents an early attempt to introduce spiking-based attention for multimodal sensor fusion in autonomous driving~\cite{ghoreishee2025}. The framework employs the standard spiking attention formulation introduced in \cite{zhou2022spikformer} to fuse vision-based sensory modalities, and further proposes a more energy-efficient spiking cross-attention module that leverages ternary neurons to compute attention scores. Although the study successfully demonstrates that multi-modal information fusion within a spiking Q-learning pipeline is feasible, the average reward achieved by the proposed model remains approximately $18 \%$ lower than its non-spiking counterpart in the highway scenario, indicating that substantial performance gaps still persist.

In general, existing vision-based spiking deep Q-learning approaches rely heavily on rate coding or are limited to low-dimensional inputs, and no standardized, computationally efficient encoding framework currently exists. Moreover, these methods typically produce quantized Q-values due to the discrete nature of spikes, making accurate value approximation challenging. We hypothesize that these limitations jointly contribute to the performance gap between spiking and non-spiking deep Q-networks. Closing this gap is critical for paving the way toward spiking-based reinforcement learning systems capable of real-time decision-making and control in safety-critical domains such as autonomous driving.

\section{Preliminaries}\label{sec:preliminaries}
The autonomous driving problem is formulated as a Markov Decision Process $(\mathcal{S}, \mathcal{A}, P, r, \gamma)$, where $\mathcal{S}$ denotes the state space, $\mathcal{A}$ the action space, $P(s' \mid s, a)$ the transition dynamics, $r(s,a)$ the reward function and $\gamma \in (0,1)$ the discount factor. At each time step $t$, the agent receives a multi-modal observation $o_t = \{o_t^{\text{cam}},\, o_t^{\text{lidar}},\, o_t^{\text{imu}}\}$, selects an action $a_t \in \mathcal{A}$, and aims to maximize the expected discounted return $\mathbb{E}[\sum_{t=0}^{T} \gamma^t r_t]$.

Sensory inputs consist of camera frames $o_t^{\text{cam}}$, LiDAR point clouds $o_t^{\text{lidar}}$, and IMU measurements $o_t^{\text{imu}}$ that provide the velocity and orientation of the ego-vehicle. Camera and LiDAR observations are transformed into spatially aligned bird’s-eye-view (BEV) representations using ego-motion compensation derived from the IMU signal, resulting in the network input $I_t = \{I_t^{\text{cam}},\, I_t^{\text{lidar}}\}$, where $I_t^{\text{cam}}$ and $I_t^{\text{lidar}}$ denote the BEV-encoded camera and LiDAR modalities, respectively.

The MM-DSQN developed by Ghoreishee et al. selects actions, guided by multi-sensory visual data~\cite{ghoreishee2025}. We now briefly describe this network, shown in Fig.~\ref{fig1}, since it provides the starting point for our solution. The MM--DSQN accepts images $I_1, I_2 \in \mathbb{R}^{C \times H \times W}$ obtained using two sensory modalities, where $C$ is the number of channels and $H$ and $W$ are the heights and widths of the images, respectively. Although the native resolutions of the images may differ, a unified notation is adopted for the sake of clarity. The channel dimension may correspond to RGB images or stacked frame histories~\cite{mnih2015human}. The two modalities consist of a camera-based BEV representation obtained via geometric projection or learned depth estimation~\cite{li2024bevformer, philion2020lift}, and LiDAR point clouds projected into pseudo-images encoding attributes such as height, intensity, and relative velocity in pixel values~\cite{lang2019pointpillars}. First, both inputs are converted to spike trains using a rate encoder $\mathcal{E}_{\mathrm{enc}}$ as 
\begin{equation*}
    X_1 = \mathcal{E}_{\mathrm{enc}}(I_1), \; 
    X_2 = \mathcal{E}_{\mathrm{enc}}(I_2), \; X_1, X_2 \in \{0,1\}^{T \times C \times H \times W}
\end{equation*}
where $T$ is the number of time steps. To reduce the computational cost of performing cross-attention over pixels in full-resolution images, the spike trains are passed through convolutional spiking layers to extract low-dimensional features 
\begin{align*}
    X_1' &= \mathrm{CONV}(X_1) \nonumber \\ 
    X_2' &= \mathrm{CONV}(X_2), \; \quad X_1',X_2' \in \{0,1\}^{T \times c \times h \times w}
\end{align*}
where $h < H, w < W$ and $c > C$. Each $\mathrm{CONV}(\cdot)$ block consists of stacked convolutions followed by leaky integrate-and-fire neurons to produce compact spiking feature maps. Each spatial location in $X_1'$ and $X_2'$ is treated as a token of dimension $c$ and projected through feature-embedding layers
\begin{equation*}
    E_{I_1} = \mathrm{EMB}(X_1') \quad \mathrm{and} \quad E_{I_2} = \mathrm{EMB}(X_2'),
\end{equation*}
where $E_{I_1}, E_{I_2} \in \{0,1\}^{T \times h \times w \times c}$. The resulting feature maps are fused through a Transformer-style cross-fusion layer, denoted $\mathrm{CFL}(\cdot)$, which incorporates the ternary spiking cross-attention mechanism~\cite{ghoreishee2025}. This module consists of cross-attention blocks and feed-forward sublayers, each equipped with residual connections and layer normalization, enabling effective multi-modal feature alignment and interaction.
\begin{equation*}
    F_{I_1,I_2} = \mathrm{CFL}(E_{I_1},E_{I_2}), \; \mathrm{where} \;
    F_{I_1,I_2} \in \{0,1\}^{T \times h \times w \times c}.
\end{equation*}
Finally, $F_{I_1,I_2}$ is reshaped to $\{0,1\}^{T \times c \times h \times w}$, flattened and passed through fully connected layers to produce the MM--DSQN output as 
\begin{equation*}
    r_a = \mathrm{FC}\big(\mathrm{flatten}(F_{X_1,X_2})\big).
\end{equation*}
The decoder then maps $q_s$ to continuous action values as $Q(a \mid I_1, I_2) = \mathcal{E}_{\mathrm{dec}}(r_a). \quad a \in \mathcal{A}$.

Two fundamental limitations hinder the effectiveness of MM--DSQNs in vision-based autonomous driving.

When rate-based coding is used to convert continuous sensory observations into spike trains, the MM–DSQN requires long simulation windows to obtain reliable firing-rate estimates, which introduces significant latency, increases energy consumption, and reduces perceptual fidelity for high-dimensional visual inputs. Such latency is incompatible with autonomous driving, which demands high decision frequencies (around 10 Hz). Consequently, the simulation time window must be kept short, further degrading the perceptual fidelity of the rate-based encoder. 

The neurons comprising the output layer of the MM-DSQN fire sparsely, and the weighted-sum decoding strategy, such as the one used by Ghoreishee et al., therefore, produces \emph{sparse Q-value estimates}, where many action values collapse towards similar magnitude~\cite{ghoreishee2025}. When Q-values cluster in this way, the agent cannot reliably discriminate between actions, and in extreme cases, where only a single action in a given state yields a nonzero reward, the remaining Q-values remain near a similar value throughout training. Such sparsity leads to poor value-function approximation, biased temporal-difference updates, and sub-optimal policies, which are especially harmful in dense and rapidly changing driving environments that require fine-grained action discrimination.

\section{Technical Approach}\label{sec:approach}
We hypothesize that the fundamental representational bottlenecks of vision-based MM--DSQNs described above can be mitigated through a computationally efficient encoding--decoding strategy expressive enough to increase the perceptual fidelity and prevent the sparse Q-value problem. Toward this end, we introduce a population-based fuzzy encoder and decoder approach that can be easily integrated within the MM--DSQN architecture, described previously in Section~\ref{sec:preliminaries}.

\begin{figure}[t!]
    \centering

    \begin{subfigure}{0.99\linewidth}
        \centering
        \includegraphics[width=\linewidth]{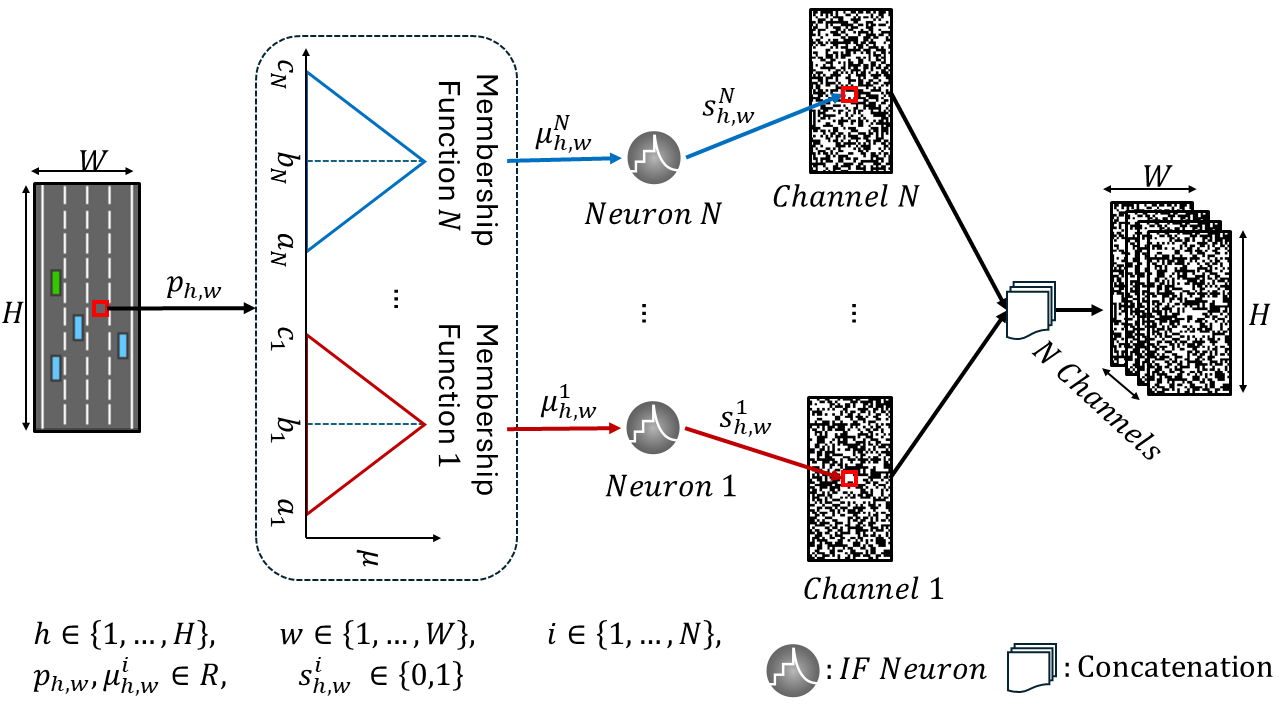}
        \caption{Population-based Encoder}
        \label{fig:encoder}
    \end{subfigure}

    \vspace{2mm}

    \begin{subfigure}{0.99\linewidth}
        \centering
        \includegraphics[width=\linewidth]{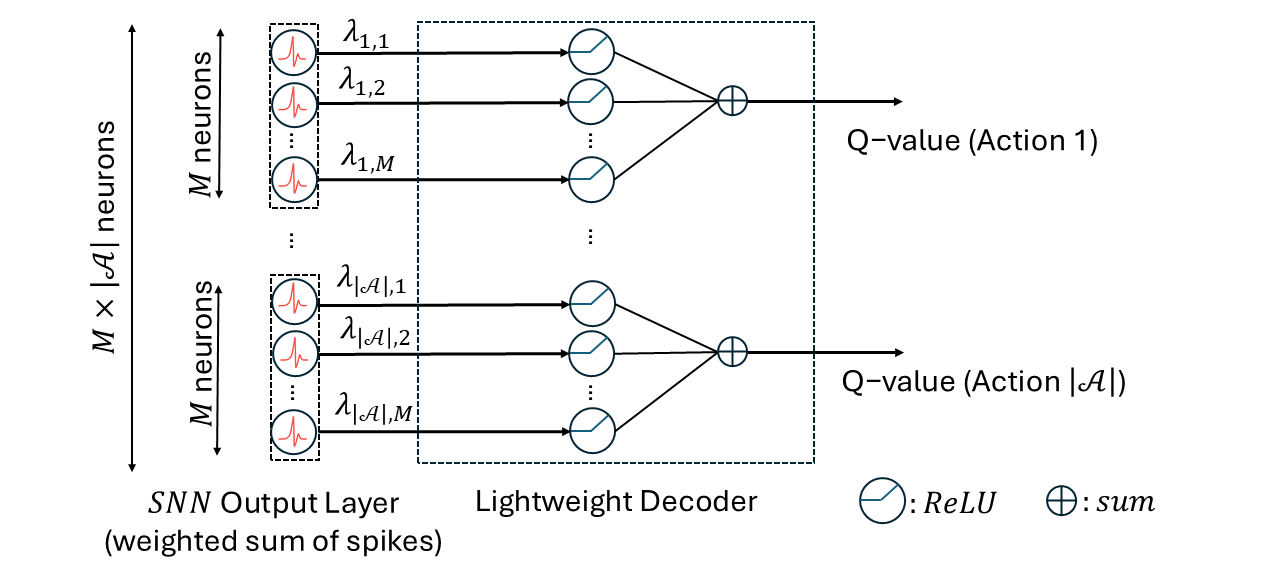}
        \caption{Lightweight Decoder}
        \label{fig:decoder}
    \end{subfigure}

    \caption{Population-based encoder using $N$ triangular fuzzy membership functions and lightweight neural decoder that uses $|\mathcal{A}|$ populations of $M$ neurons.}
    \vspace{-12pt}
    \label{fig2}
\end{figure}

\subsection{Design of the Encoder}
The encoder, shown in Fig.~\ref{fig:encoder}, transforms each pixel 
$p_{h,w}$ of an input image channel $I \in \mathbb{R}^{H \times W}$ into a 
higher-dimensional fuzzy population representation. Each pixel is processed by 
a bank of $N$ trainable fuzzy membership functions, resulting in $N$ activation 
values in $[0,1]$. These activations are then fed into $N$ integrate-and-fire 
(IF) neurons, each producing a spike train that encodes the corresponding 
degree of membership. As a result, every original image channel is expanded into 
$N$ distinct spiking channels, and concatenating these channels across all 
pixels yields the final spike-encoded image.

For computational efficiency—particularly with high-resolution inputs—the 
encoder uses \textit{triangular} membership functions, which rely solely on 
simple linear operations and, therefore, remain lightweight and suitable for 
real-time processing. Each membership function is defined as
\begin{equation*}
\mu^i(p_{h,w};a_i,b_i,c_i) =
\begin{cases}
0, & p_{h,w} \le a_i,\\[4pt]
\dfrac{p_{h,w}-a_i}{\,b_i-a_i\,}, & a_i < p_{h,w} \le b_i,\\[6pt]
\dfrac{c_i-p_{h,w}}{\,c_i-b_i\,}, & b_i < p_{h,w} \le c_i,\\[6pt]
0, & p_{h,w} > c_i,
\end{cases}
\end{equation*}
where $a_i$, $b_i$, and $c_i$ are the trainable parameters of the $i$-th fuzzy 
membership function, and $\mu^i(p_{h,w})$ denotes the corresponding membership 
degree.

A pixel value is passed through all $N$ membership functions, generating $N$ membership degrees that are subsequently converted into spike trains by the associated IF neurons. These spike trains collectively form the encoded representation of that pixel across 
the $N$ new spiking channels. To illustrate the encoding behavior, consider two pixel values, $p_1 = 0.35$ and 
$p_2 = 0.75$, evaluated under three triangular membership functions with 
parameters:
\begin{align*}
    (a_1,b_1,c_1) &= (0.0,0.2,0.4), \\
    (a_2,b_2,c_2) &=(0.3,0.5,0.7), \\ 
    (a_3,b_3,c_3)&= (0.6,0.8,1.0).
\end{align*}
For pixel value $p_1 = 0.35$, the activations are
\begin{equation*}
    \mu^1(p_1) = 0.25, \; \mu^2(p_1) = 0.25, \; \mbox{and} \; \mu^3(p_1) = 0,
\end{equation*}
while for pixel value $p_2 = 0.75$, the activations are
\begin{equation*}
    \mu^1(p_2)=0,\;  \mu^2(p_2)=0, \; \mbox{and} ;\ \mu^3(p_2)=0.75.
\end{equation*}
Therefore, low pixel values activate the first two channels, whereas higher 
values strongly activate the third. \emph{By mapping a scalar pixel intensity into a 
distributed channel response, the encoder enriches the representational capacity 
provided to the downstream SNN, enabling finer discrimination of image 
features and more expressive spiking dynamics.}

\subsection{Design of the Decoder}
We allocate a population of $M$ output neurons to each action, which 
yields $M \times |\mathcal{A}|$ neurons in the SNN output layer, where 
$|\mathcal{A}|$ denotes the number of discrete actions. For each action 
$a \in \mathcal{A}$, the associated population produces $\{\lambda_{a,1}, \lambda_{a,2}, \ldots, \lambda_{a,M}\}$, 
which serves as a population-based fuzzy encoding of its Q-value 
(Fig.~\ref{fig:decoder}). Each element $\lambda_{a,m}$ is calculated as
\begin{equation*}
    \lambda_{a,m}
    = \sum_{t=1}^{T} \sum_{i} w_i\, s_{i,t},
    \label{eq:spikeacc}
\end{equation*}
where $s_{i,t} \in \{0,1\}$ denotes the spike emitted by the hidden-layer neuron $i$ 
at time $t$, and $w_i$ is the synaptic weight connecting the spiking neuron $i$ to
the $(a,m)$-th output unit.

Classical fuzzy systems aggregate these activations into a membership function
and apply a defuzzification operator---such as the centroid, bisector, or
weighted average. For instance, the centroid operator yields
\begin{equation*}
    Q(a)
    =
    \frac{\displaystyle \int x\, \mu_a(x)\, dx}
         {\displaystyle \int \mu_a(x)\, dx},
    \label{eq:centroid}
\end{equation*}
where $\mu_a(x)$ denotes the aggregated membership function for action $a$. 
Although theoretically grounded, such defuzzification operators are computationally demanding 
and scale poorly in reinforcement-learning settings with large action spaces.

To eliminate this overhead, we introduce a lightweight neural decoder shown in Fig.~\ref{fig:decoder} that learns an efficient mapping from the fuzzy population activations to continuous Q-values. The decoder is implemented as a 
compact fully connected network with ReLU nonlinearities, providing a fast, 
differentiable, and hardware-efficient approximation of the defuzzification process.

\subsection{End-to-End Training}
Assume two observations $I_1, I_2 \in \mathbb{R}^{C \times H \times W}$. 
The proposed fuzzy encoder--SNN--decoder pipeline is formulated as
\begin{align*}
    & X_1 = \mathcal{E}_{\mathrm{fuz\text{-}enc}}(I_1),
    \qquad
    X_2 = \mathcal{E}_{\mathrm{fuz\text{-}enc}}(I_2) \\
    & \lambda_a = \{ \lambda_{a,1}, \lambda_{a,2}, \ldots, \lambda_{a,M} \}
    = \mathcal{SNN}(X_1, X_2),
    \qquad a \in \mathcal{A} \\
    & Q(a \mid I_1, I_2)
    = \mathcal{E}_{\mathrm{fuz\text{-}dec}}(\mathbf{\lambda}_a),
    \qquad a \in \mathcal{A}
\end{align*}
Here, $\mathcal{E}_{\mathrm{fuz\text{-}enc}}$ denotes the population-based fuzzy encoder, $\mathcal{SNN}$ is the multimodal deep spiking Q-network,  
and $\mathcal{E}_{\mathrm{fuz\text{-}dec}}$ represents the lightweight fuzzy decoder.
Both the fuzzy encoder and decoder are trained jointly with the spiking Q-network.
All parameters, including the membership-function parameters, are optimized
end-to-end via backpropagation, enabling the model to learn a task-specific
fuzzy representation space that reduces information loss during spike encoding
and improves Q-value estimation.
 
\section{Theoretical Analysis}\label{sec:analysis}
This section provides an analysis of the information gain achieved by our approach and the incurred computational cost.

\subsection{Information Gain}
We quantify the additional representation capacity due to the encoder and decoder within the MM-DSQN by analyzing the available entropy in the network. Recall that the maximum entropy of a discrete set $Z$ is 
\begin{equation*}
    C(Z) =  \max \Bigl( - \sum_{z \in \mathcal{Z}} p_Z(z)\, \log p_Z(z) \Bigr),
\end{equation*}
where $p_Z(z)$ is the probability of observing element $z \in \mathcal{Z}$. The quantity $C(Z)$, which is the information capacity, is maximized when $Z$ follows a uniform distribution, i.e., $p_Z(z) = 1/N$ for $N = |Z|$. Under this condition, $C(Z) = \log N$, which allows for a direct comparison between the representational capacity of: (i) the raw non-encoded input, (ii) classical rate-based spike encoding, and  (iii) population-based fuzzy encoding.

For \emph{non-encoded input} that uses 32-bit floating-point tensors $I \in \mathbb{R}^{C\times H\times W}$, the maximum
entropy is
\begin{equation*}
    C(I) = \log \!\left( 2^{32\,C\,H\,W} \right)
          = 32\,C\,H\,W .
\end{equation*}

For \emph{rate-based encoding}, let $X_{\mathrm{rate}} \in \{0,1\}^{T \times C \times H \times W}$ denote the 
rate-coded spike trains obtained from $I$ using a 1-bit representation over $T$  simulation steps. Its representational capacity is
\begin{equation*}
    C(X_{\mathrm{rate}})
    = \log\!\left( 2^{T\,C\,H\,W} \right)
    = T\,C\,H\,W .
\end{equation*}
For small $T$, which is essential in real-time systems such as autonomous 
driving, the information loss relative to 32-bit inputs is substantial.

Finally, for \emph{fuzzy encoding}, let $X_{\mathrm{pop}} \in \{0,1\}^{T \times N C \times H \times W}$ be the 
population-coded version of $I$, where each pixel is represented by a population of 
$N$ fuzzy-encoded neurons.  
The representational capacity becomes
\begin{equation*}
    C(X_{\mathrm{pop}})
    = \log\!\left( 2^{N\,T\,C\,H\,W} \right)
    = N\,T\,C\,H\,W .
\end{equation*}
Thus, even with small $T$, the multiplicative factor $N$ significantly mitigates 
information loss and enables richer spike-based representations. A similar effect occurs during decoding. A single real-valued Q-value stored as a 32-bit floating-point number has entropy $C(Q) = 32$. A single spike train provides at most $T$ bits of information. Using a population of $M$ fuzzy-decoding neurons increases this upper bound to $C(Q_{\mathrm{pop}}) = M\,T$, providing a substantially larger representational space to reconstruct continuous  Q-values. 

Overall, above analysis shows that classical rate-based encoding severely constrains 
representational capacity when the simulation window is short, whereas the proposed 
population-based fuzzy encoder--decoder substantially reduces information loss in both 
the encoding and decoding stages.

\subsection{Computational Overhead}
When using \emph{non-encoded input} $x \in \mathbb{R}^{c \times h \times w}$, the first convolutional layer dominates the computation. With $c_{\text{out}}$ kernels of size $l \times l$, stride $s$, and padding $p$, the total number of multiplications is $c_{\text{out}}\, c\, l^{2}\, h_{\text{out}}\, w_{\text{out}}$,
where $h_{\text{out}} = \Big\lfloor \frac{h + 2p - l}{s} \Big\rfloor + 1$ and $w_{\text{out}} = \Big\lfloor \frac{w + 2p - l}{s} \Big\rfloor + 1$. Modern CNNs typically use large $c_{\text{out}}$ and $l$, making this configuration the most computationally expensive among the three encoding strategies.

Rate encoding does not perform multiplicative operations per pixel, while in the fuzzy encoder, each pixel is processed by a population of $N$ triangular membership functions, requiring only a small and constant number of arithmetic operations per pixel. The multiplication count is $c\, N\, h\, w$. Since normalized pixel values lie in $[0,1]$, a small population (for example, $N=3$) is generally sufficient, resulting in substantially fewer operations than the non-encoded case while providing richer representational capacity than rate encoding.

Overall, rate encoding incurs the lowest computational overhead, the non-encoded case the highest, and population-based fuzzy encoding offers a tunable trade-off between computational efficiency and representational fidelity.

The decoding layer incurs additional overhead due to the small neural layer, which can be quantified as $M \times |\mathcal{A}|$, where $|\mathcal{A}|$ is the number of discrete actions and $M$ is the population size at the decoder. In high-level decision-making tasks, $|\mathcal{A}|$ is typically small and $M$ is a tunable parameter (usually $1$--$20$). Thus, the computational overhead of the decoder remains negligible in relation to the overall cost of the network.

\section{Performance Evaluation}\label{sec:evaluation}
This section evaluates the encoding--decoding architecture using an autonomous driving scenario. We first describe the experimental setup, followed by the baseline model, implementation details, evaluation metrics, and quantitative and qualitative results. In addition, we provide ablation studies to isolate the contributions of each architectural component \footnote{The implementation is accessible through the repository at  https://github.com/Aref7792/Fuzzy-Encoding-Decodin-MM-DSQN}.

\subsection{Experimental Setup}
We evaluate the proposed architecture on a high-level decision-making task from
the \texttt{HighwayEnv} benchmark, focusing on the \textit{Highway} scenario.
\texttt{HighwayEnv} is well suited for this study because it provides a clean,
controlled, and reproducible environment for generating multimodal observations.
This allows us to isolate and assess the contribution of the proposed model
without the confounding effects introduced by high-fidelity perception stacks in
more complex simulators such as CARLA. The \textit{Highway} task requires
lane-keeping, speed regulation, and safe maneuvering under moderate traffic
conditions. The environment supplies a structured observation space comprising
three sensing modalities: kinematic variables, LiDAR measurements, and BEV
visual observations. The available action space is specified as a discrete \textit{meta-action} set that abstracts high-level driving intentions rather than low-level continuous actuation, given by
\begin{equation*}
    \mathcal{A} = \{\text{LEFT},\; \text{IDLE},\; \text{RIGHT},\; \text{FASTER},\; \text{SLOWER}\}.
\end{equation*}

The agent receives a multi-modal observation composed of three complementary sensory streams. 
\begin{itemize}
    \item \textit{LiDAR observations}: The $360^\circ$ environment is divided into fixed angular sectors, each returning two measurements: the distance to the nearest object and its relative radial velocity.

    \item \textit{BEV images}: A top-down visual representation of the surrounding scene, capturing nearby vehicles and lane geometries.

    \item \textit{Kinematic data}: Ego-vehicle states, analogous to IMU outputs: longitudinal velocity and heading angle.
\end{itemize}

The LiDAR measurements, together with the kinematic states, are transformed into an image-like occupancy grid in which each pixel denotes the spatial location of detected objects and its intensity encodes their relative velocity. These processed LiDAR---kinematic features and the BEV frames form two separate modality streams, and each is independently passed through its designated encoder variant before being processed by the MM-DSQN.

Each encoder uses three triangular fuzzy membership functions with learnable parameters, projecting every input channel into a higher-dimensional representation of three membership activation maps. For estimation of the action-value, we use a population of five neurons per action. Each neuronal population is passed through a lightweight decoding network consisting of a single hidden layer with ReLU activation, producing the final Q-value estimate for each action.

\subsection{Baseline Models}
To evaluate the contribution of the encoding-decoding approach, it is compared against two representative baselines:
\begin{itemize}
    \item Non-spiking multi-modal DQN that has been used as a baseline in Ghoreishee et al.~\cite{ghoreishee2025}.

    \item A spiking variant that uses rate encoding to convert input modalities into binary spike trains over a fixed simulation window~\cite{ghoreishee2025}. 
\end{itemize}

\subsection{Implementation Details}
All inputs first pass through a fuzzy encoder with three learnable triangular membership functions per channel. The encoded BEV and LiDAR observations are processed by three Conv2D layers (channels $8{\rightarrow}16{\rightarrow}16$) with modality-specific kernel sizes and binary LIF activation, and then projected to a 32-dimensional embedding. The fused representation is produced by a cross-attention module with eight heads, a 128-dimensional feed-forward layer, learnable positional encoding, and LayerNorm. The decision head consists of a 512-unit binary-LIF layer generating a 25-dimensional population code, which a lightweight decoder maps to five Q-values using a single ReLU-activated linear layer. Training follows standard DQN with discount factor $\gamma=0.99$, Adam optimizer with learning rate $1\times10^{-4}$, batch size 64, replay buffer size $5\times 10^{4}$, and target updates every 200 steps. Spiking dynamics uses a membrane time constant $\tau_m=2$, asymmetric thresholds $(1.0,-4)$, subtractive reset, arctangent surrogate gradients, and a simulation window of $T_s=5$.

\begin{figure}[!t]
    \centering

    \begin{subfigure}{0.9\linewidth}
        \centering
        \includegraphics[width=\linewidth]{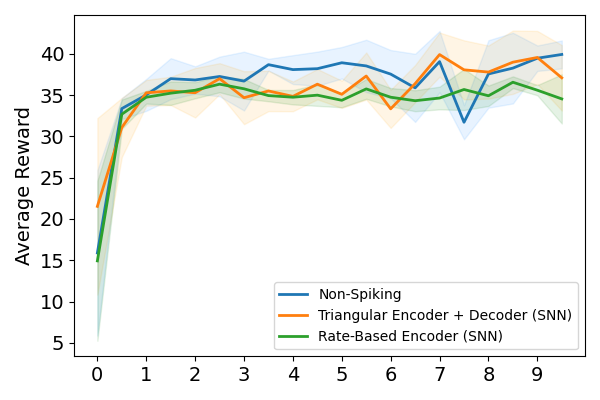}
        \label{fig:sub_reward}
    \end{subfigure}

    \vspace{-18pt}

    \begin{subfigure}{0.9\linewidth}
        \centering
        \includegraphics[width=\linewidth]{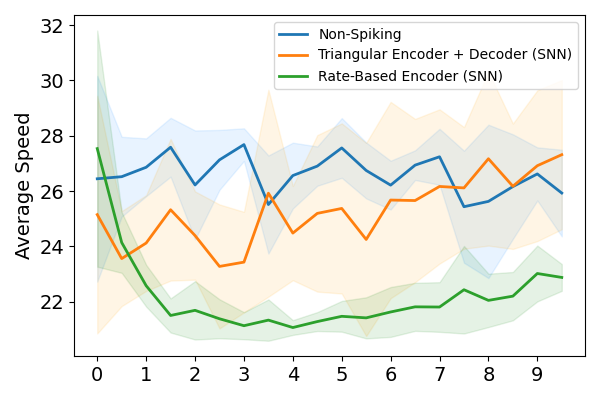}
        \label{fig:sub_speed}
    \end{subfigure}

    \vspace{-18pt}

    \begin{subfigure}{0.9\linewidth}
        \centering
        \includegraphics[width=\linewidth]{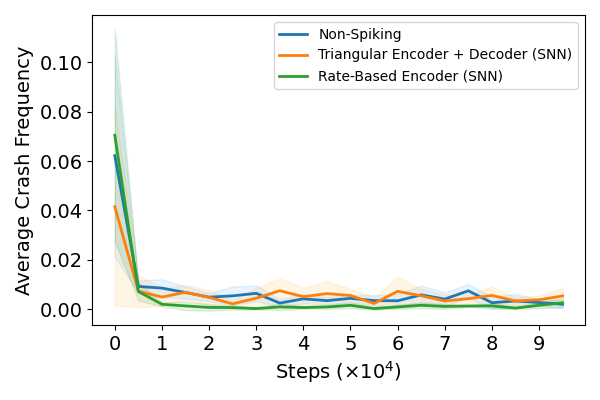}
        \label{fig:sub_crash}
    \end{subfigure}

    \vspace{-12pt}
    
    \caption{Comparison of non-spiking MM-DQN, rate-based encoding MM-DSQN, and the encoding--decoding MM-DSQN models in terms of average reward, speed, and crash frequency.}
    \label{fig:summary_metrics}
    \vspace{-12pt}
\end{figure}

\subsection{Evaluation Metrics}
All the architectures are trained with five random seeds. During training, we periodically save the policy every $5000$ steps and evaluate each saved checkpoint over 20 independent test environments. We report three metrics: (i) the average cumulative reward, (ii) the average crash frequency, and (iii) the average speed of the ego-vehicle. These metrics collectively capture decision quality, safety, and driving efficiency.
 
\subsection{Quantitative Results}
Figure~\ref{fig:summary_metrics} demonstrates a clear distinction between the behaviors of the competing models. The rate-based MM-DSQN consistently converges to an overly conservative and ultimately suboptimal policy, as reflected by its markedly lower average speed and cumulative reward relative to the non-spiking baseline. This variant persistently over-selects the \textit{SLOWER} action, revealing a systematic failure to assign appropriate Q-values to high-risk, high-reward actions such as \textit{FASTER}. Consequently, it suffers a substantial performance degradation of roughly $12.5\%$ in average reward. By contrast, the proposed encoding--decoding architecture completely eliminates this degradation: its learned policies closely match the non-spiking model in both reward and speed, particularly in the later stages of training, demonstrating a significantly more accurate estimation of Q-values for challenging high-reward decisions.

Safety metrics further underscore this improvement. The rate-based model achieves the lowest crash frequency, but at the cost of very cautious driving and reduced task performance. In contrast, both the non-spiking baseline and the encoding--decoding MM-DSQN variant maintain substantially higher speeds while still preserving safe behavior---achieving average crash frequencies of approximately $0.002$ and $0.0025$, respectively, compared to $0.001$ for the rate-based model. This demonstrates that our approach almost achieves the non-spiking performance metrics.

\begin{figure}[t!]
    \centering
    \includegraphics[width=0.9\linewidth]{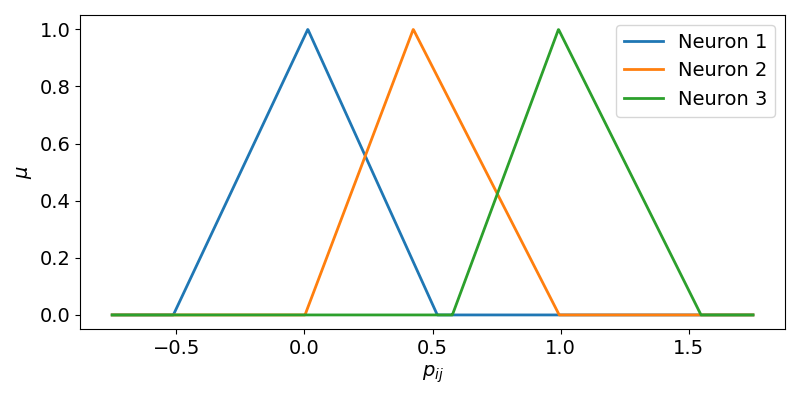}
    \caption{Learned fuzzy triangular membership functions showing how each pixel is mapped into three intensity-dependent membership channels.}
    \label{fig:fuzzy_membership}
    \vspace{-12pt}
\end{figure}

\begin{figure}[t!]
    \centering

    \begin{subfigure}{0.9\linewidth}
        \centering
        \includegraphics[width=\linewidth]{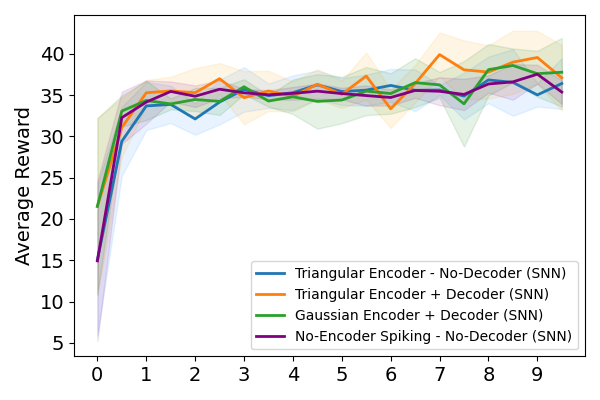}
        \label{fig:sub_reward}
    \end{subfigure}

    \vspace{-18pt}

    \begin{subfigure}{0.9\linewidth}
        \centering
        \includegraphics[width=\linewidth]{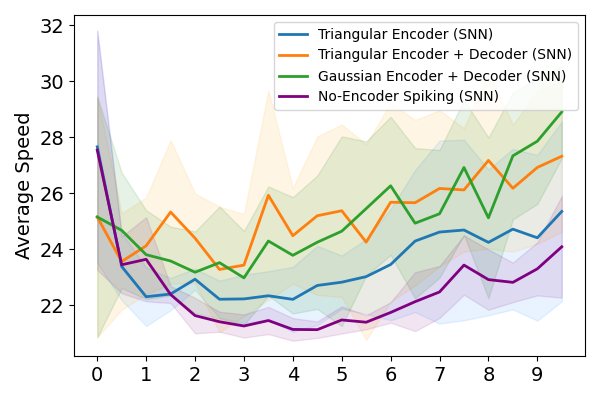}
        \label{fig:sub_speed}
    \end{subfigure}

    \vspace{-18pt}

    \begin{subfigure}{0.9\linewidth}
        \centering
        \includegraphics[width=\linewidth]{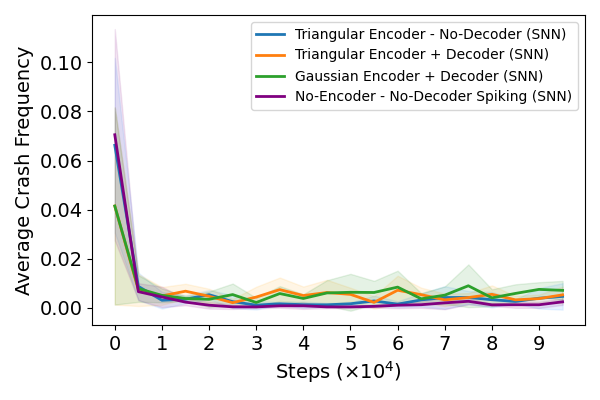}
        \label{fig:sub_crash}
    \end{subfigure}

     \vspace{-12pt}
     
    \caption{Effects of removing the decoder, removing both encoder and decoder, and replacing triangular membership functions with Gaussian ones. The triangular encoder with decoder achieves highest reward.}
    \label{fig:ablation_study}
    \vspace{-12pt}
\end{figure}

Figure~\ref{fig:fuzzy_membership} shows the learned fuzzy membership functions that drive our input encoding mechanism. The encoder transforms each input pixel into three membership activations, effectively distributing the value of the pixel across three interpretable channels: low-intensity pixels strongly activate the first channel, high-intensity pixels map predominantly to the third, and intermediate values are captured smoothly by the middle channel. This structured encoding is far more expressive than raw intensity values; it exposes continuous pixel variations to the SNN in a simpler to discriminate form, stabilizes the downstream Q-value estimation process, and provides the SNN with a more informative view of the environment.

\subsection{Ablation Study}

Figure~\ref{fig:ablation_study} quantifies the contribution of each architectural component. Removing only the decoder and replacing it with a simple weighted sum of spikes produces a clear degradation in reward, speed, and crash frequency, confirming that the decoder is essential for accurate estimation of Q-values. Eliminating both the encoder and decoder forces the first layer to operate using floating-point values rather than spikes, effectively making it non-spiking; however, this configuration still performs worse than the proposed triangular encoder with decoder, underscoring the importance of the decoding stage. We also evaluate the use of Gaussian membership functions in place of triangular ones. Although Gaussian functions incur a higher computational cost due to exponential operation, the triangular encoder consistently yields superior performance, achieving higher reward and lower crash frequency overall.

\section{Conclusions}\label{sec:conclusions}
We have developed a light-weight fuzzy encoding and decoding architecture for vision-based MM-DSQNs that learns optimal fuzzy membership functions to minimize information loss during spike encoding and uses a population-based decoder to reconstruct expressive Q-values from sparse spike outputs. For the Highway scenario in the Highway-Env benchmark, we have demonstrated that augmenting an MM-DSQN using this architecture allows the spiking model to match the performance of its non-spiking counterpart. 

In future work, we will evaluate performance in more complex driving scenarios (roundabouts and intersections). We also plan to apply the augmented MM-DSQNs in RL-tasks in multi-agent scenarios.  

\section{Acknowledgments}
This material is based upon work supported by the National
Science Foundation under Grant No 2209745.

\printbibliography

\end{document}